\documentclass[lettersize,journal]{IEEEtran}
\usepackage{amsmath,amsfonts}
\usepackage{array}
\usepackage[caption=false,font=normalsize,labelfont=sf,textfont=sf]{subfig}
\usepackage{textcomp}
\usepackage{stfloats}
\usepackage{url}
\usepackage{verbatim}

\usepackage{graphicx}
\usepackage{cite}
\usepackage{multirow}
\usepackage{graphicx}

\usepackage{bbm}
\usepackage{bm}
\usepackage{threeparttable}
\usepackage[ruled,linesnumbered,boxed]{algorithm2e}

\begin{document}

\title{TrajGEOS: Trajectory Graph Enhanced Orientation-based Sequential Network for Mobility Prediction}

\author{Zhaoping Hu{$^+$},~Zongyuan Huang{$^+$},~Jinming Yang,~Tao Yang,~Yaohui Jin{$^\ast$},~\IEEEmembership{Member,~IEEE},~Yanyan Xu{$^\ast$}
\IEEEcompsocitemizethanks{
\IEEEcompsocthanksitem Z. Hu, Z. Huang, J. Yang, Y. Jin, Y. Xu are with the MoE Key Laboratory of Artificial Intelligence and AI Institute, Shanghai Jiao Tong University, Shanghai 200240, China. E-mail: \{zhaopinghu, herozen, yangjm67, jinyh, yanyanxu\}@sjtu.edu.cn. 
\IEEEcompsocthanksitem Tao Yang is with Shanghai Transportation Information Center, Shanghai Urban-Rural Construction and Transportation Development Research Institute, Shanghai 200032, China. E-mail: yangtaocoolboy@163.com\\
$^+$ Equal contribution. \\
$^\ast$ Corresponding \ authors. \\

}% <-this % stops a space
}

\maketitle

\begin{abstract}
Human mobility studies how people move to access their needed resources and plays a significant role in urban planning and location-based services. As a paramount task of human mobility modeling, next location prediction is challenging because of the diversity of users' historical trajectories that gives rise to complex mobility patterns and various contexts. Deep sequential models have been widely used to predict the next location by leveraging the inherent sequentiality of trajectory data. However, they do not fully leverage the relationship between locations and fail to capture users' multi-level preferences. This work constructs a trajectory graph from users' historical traces and proposes a \textbf{Traj}ectory \textbf{G}raph \textbf{E}nhanced \textbf{O}rientation-based \textbf{S}equential network (TrajGEOS) for next-location prediction tasks. TrajGEOS introduces hierarchical graph convolution to capture location and user embeddings. Such embeddings consider not only the contextual feature of locations but also the relation between them, and serve as additional features in downstream modules. In addition, we design an orientation-based module to learn users' mid-term preferences from sequential modeling modules and their recent trajectories. Extensive experiments on three real-world LBSN datasets corroborate the value of graph and orientation-based modules and demonstrate that TrajGEOS outperforms the state-of-the-art methods on the next location prediction task.
\end{abstract}


\section{Introduction}

Human mobility prediction is of great importance to numerous applications, including traffic scheduling~\cite{xu2017collective, ccolak2016understanding}, electric vehicle charging management~\cite{xu2018planning}, urban planning~\cite{barbour2019planning, xu2021emergence}, and decision-making during large-scale emergency events~\cite{yabe2022mobile}. Due to the deep-rooted regularity of people's daily mobility~\cite{song2010limits}, mobility modeling is scientifically possible, and numerous research fields have emerged in the past decade~\cite{xu2021understanding}. Besides, the increasing prevalence of mobile phones, GPS services, and digital maps contribute to the surge of location-based social networks (LBSNs) such as Foursquare, Gowalla, Yelp, etc. User-generated content in LBSNs contains not only spatial and temporal stamps of activity but also contextual information. As the explosive growth of LBSNs made the activity data of millions of users~\cite{yang2014modeling} more accessible, it also brought great convenience to human mobility prediction tasks, especially for next location prediction.

As an important task of human mobility modeling~\cite{tang2022hgarn}, next-location prediction aims to predict the most possible location an individual user is going to visit. 
Most current research utilizes sequential models to capture users' behavioral patterns and deduce their preferences, due to the inherent sequentiality of human mobility actions. 
These conventional approaches tend to employ vanilla sequential models, compounded with augmented spatial-temporal features, such as distance-matrix, to investigate behavioral patterns from massive historical trajectories. With the advancement of deep learning, recurrent neural networks (RNNs) have gained extensive acceptance for capturing sequential dependencies in next-location prediction tasks.
To further make full use of spatial-temporal information, researchers have designed novel gate mechanisms~\cite{zhao2020go} and introduced additional spatial-temporal features~\cite{liu2016predicting, yang2020location, kong2018hst, manotumruksa2018contextual, chen2020context} to RNNs. Advanced techniques such as attention mechanism~\cite{feng2018deepmove, luo2021stan} and transformer~\cite{yang2022getnext, hong2022you, xue2021mobtcast} were also applied to users' preference modeling with their historical sequences. 

\begin{figure}[htbp]
  \centering
  \includegraphics[width=0.95\linewidth]{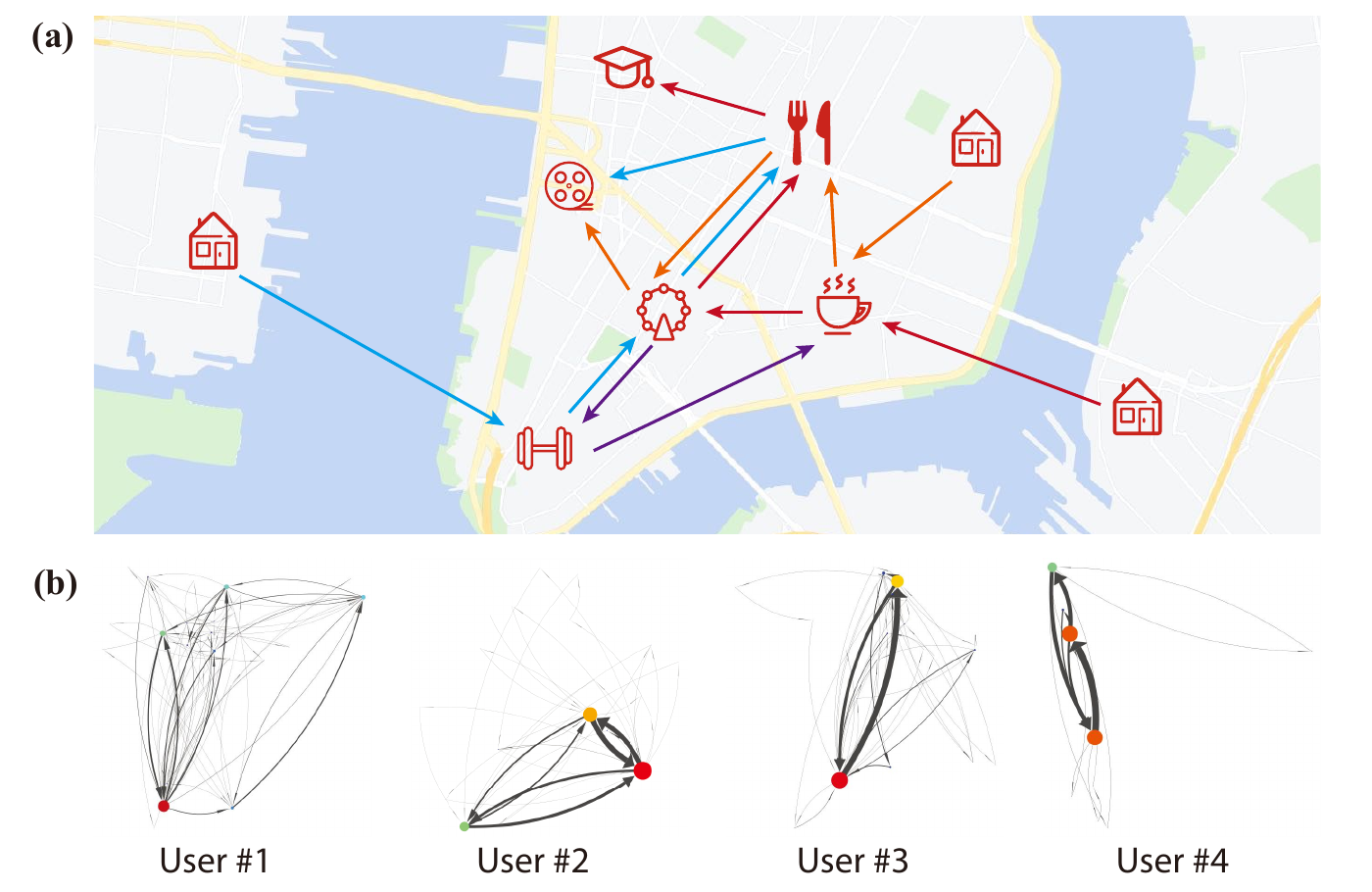}
  \caption{(a) Implicit relationship between locations. (b) Sample users' trajectory graphs built from their historical check-in data.}
  \label{fig:story}
  % \Description{ (a) There are similar transition patterns in different users' historical trajectories. (b) The user's historical trajectory can be recorded as a trajectory graph, which contains the users stable movement pattern and implicit preference. }
\end{figure}

The sequential modeling approaches provide critical insights into the next-location prediction problem, and experiments on real-world datasets confirmed their effectiveness. 
% There are, however, three intractable drawbacks to these approaches. 
There are, however, two intractable drawbacks to these approaches. 
(i) They insufficiently explore the relationship among diverse locations. 
As Figure~\ref{fig:story} (a) shows, there are many records of movements between the amusement park and the restaurant, which implies that they may be geographically close and have similar contexts.
Yet, despite most sequential models learning the behavior of individual users, they fail to learn the global relation across different locations, which is crucial for downstream prediction tasks. 
% (ii) Most sequential modeling approaches focus solely on the sequence of locations but do not take full advantage of users' activity chains, considering the fact that people's activities and visited locations probably are closely related. For example, people usually go to coffee shops to enjoy afternoon tea and go to gyms to exercise.  
% (iii) 
(ii) How to effectively incorporate all historical trajectories to model mobile behavior is another critical issue for most existing models. 
It's difficult for vanilla sequential models to capture long-term transition patterns with all historical trajectories. 

Towards these issues, in this paper, we propose TrajGEOS, a trajectory graph enhanced orientation-based sequential network for mobility prediction. 
First, to capture the global relation between locations, we construct a large trajectory graph that aggregates all historical mobility sequences of users and applies a hierarchical graph learning approach to generate location and user representations that contain spatial and contextual relations. 
These representations serve as important supplements to trajectory encoding in the sequential module. 
Our model incorporates next-visit category prediction as an auxiliary task, enabling it to access additional features that contain categorical information when modeling location trajectories. 
Besides, to better capture the user's mobility patterns and preferences, we design a next-location predictor that integrates long-term, short-term, and mid-term preferences (multi-preferences). Users' long-term preferences come from a location-weighted readout of the individual trajectory graph like Figure~\ref{fig:story} (b) shows, and short-term preferences are drawn from the trajectory sequences within the current week. For the mid-term preference, we design an orientation module that utilizes position embedding and attention mechanism based on their trajectories over the past two weeks. 
The implementation of this work will be released after publication.

Our contributions can be summarized as follows:
\begin{itemize}
    \item We design a hierarchical graph convolution approach to process the global trajectory graph. The lower convolution is applied to obtain the location embedding, while the upper convolution is applied to user-specific trajectory subgraphs to derive user representations. 
    This structure enables further exploration of both location relationships and long-term preferences among users within their historical trajectories. 
    \item We propose the TrajGEOS model, which captures users' multi-preferences, including long-, mid-, and short-term transition patterns, to leverage all historical records more effectively.  
    % todo
    % Specifically, it leverages hierarchical graph convolution on trajectory graphs to extract location representation and long-term user preferences, while also incorporating the sequential module and the orientation-based module for short- and mid-term preferences. 
    The final predictor outputs the next location based on the user's multi-preference derived from their entire historical records. 
    \item We conduct extensive experiments on three real-world location-based social networks (LBSNs) check-in datasets, and further confirm the effectiveness of our model through ablation experiments and visualization analyses.
\end{itemize}

\section{Related Work}

\subsection{Next location prediction}

Current next location prediction methods mainly focus on the temporal patterns of the historical trajectory when predicting an individual's next movement. Related approaches can be classified into two categories: traditional methods and deep learning methods.
Traditional methods are mostly pattern-based~\cite{rendle2010factorizing, cheng2013you, ye2011exploiting} or machine-learning based~\cite{cheng2012fused, li2015rank, lian2014geomf, feng2015personalized, feng2020hme, zhao2016stellar}. 
For instance, FPMC~\cite{rendle2010factorizing} adds user-personalized transition matrices and FPMC-LR~\cite{cheng2013you} adds an additional localized constraint to capture users' transition patterns. 
\cite{ye2011exploiting} developed a naive Bayesian-based method combined with geographical influence to recommend the next location. 
With the development of matrix factorization-based methods in recommender systems, some researchers incorporate additional geographical and social influence~\cite{cheng2012fused, li2015rank} or spatial clustering constrain~\cite{lian2014geomf} into the matrix factorization method to recommend the next place for users. 
The metric embedding method can better model the sequential transition with the strategy of representing each item as a single point in the latent space. So some works incorporate individuals' preference~\cite{feng2015personalized}, category, and region information~\cite{feng2020hme} to the metric embedding method and predict users' next locations. 
These traditional approaches are limited to feature engineering, which usually requires domain knowledge, making it difficult to construct the relationships between unstructured features from multi-format data.

Since the user trajectory data in LBSN naturally has a sequential structure, most deep learning methods use sequential models such as recurrent neural networks to capture mobility patterns. 
Some researchers also incorporate additional tricks, including attention mechanism~\cite{kong2018hst, wu2019long, feng2018deepmove, luo2021stan, lian2020geography} and flashback~\cite{yang2020location}, to enhance the models and make full use of data sparsity. 
~\cite{liu2016predicting, kong2018hst, sun2020go} add additional spatial-temporal influences and geographical relations to RNN, which help the model capture patterns in historical trajectory data. 
\cite{wu2019long} proposed long and short-term modules to learn preferences, and ~\cite{wu2020personalized} incorporate additional personalized weights for individualized recommendation. 
STGCN~\cite{zhao2020go} designed a new LSTM with additional gate mechanisms to capture spatial-temporal features. 
STAN~\cite{luo2021stan} and GeoSAN~\cite{lian2020geography} exploit additional geographical and temporal information to predict the next location based on self-attention networks. 
Considering the geographical impact on location prediction, DIG~\cite{qin2022disentangling} disentangles the geographical and user interest factor, utilizing a geo-constrained negative sampling strategy and soft-weighted loss function.
SSDL~\cite{gao2023predicting} disentangle time-invariant and time-varying factors in human mobility patterns, utilizing trajectory augmentation techniques to mitigate data sparsity, and incorporating a POI-centric graph structure to capture heterogeneous collaborative signals from historical check-ins. 
CSLSL~\cite{huang2024human} integrates causal structures and spatial constraints to explicitly modeling the decision logic of human mobility and ensuring consistency between predicted and actual spatial distributions.
Graph-based methods for next location prediction have garnered significant attention in recent years. These approaches leverage the structural relationships between locations, providing a powerful framework for understanding and forecasting human mobility patterns. In the following subsection, we will provide a detailed overview of these methods.

% Yet, all these approaches do not sufficiently explore the relationships across locations.

\subsection{Graph learning in mobility prediction}

Since graph convolution neural networks emerged as an innovative method to model structured graph data, graph learning approaches have attracted extensive attention and have been widely used in different tasks. 
Point of interest (POI) recommendation is a usual format of next location prediction that is closely related to the recommendation task. 
Some recent work on the next POI recommendation leverages the graph embedding method to enhance their models with geospatial information that can be used in downstream prediction tasks. 
For instance, GE~\cite{xie2016learning} uses POI-POI, POI-Region, POI-Time, and POI-Word bipartite graphs to capture different paradigms and make recommendations for users' next POI.  
DYSTAL~\cite{xiong2020dynamic} jointly learns the embedding of users and locations from three graphs, i.e., POI-POI, user-POI, and user-user, and excavates spatial-temporal patterns based on historical trajectories; also, it designs a dynamic factor graph to capture the different factors from the network embedding module. 
The GETNext model~\cite{yang2022getnext} combines graph neural network technique and transformer~\cite{vaswani2017attention} structure to predict the next location. It uses a unified graph constructed of check-in sequences to generate the embedding of locations, and this graph construction method reflects the global transition patterns of all users explicitly. 
Graph-Flashback~\cite{rao2022graph} introduces a Spatial-Temporal Knowledge Graph and integrates both spatiotemporal information and user preferences to explicitly learn weighted POI transition graphs. 
STHCN~\cite{yan2023spatio} leverages a hypergraph to capture both intra-user and inter-user trajectory information and incorporates a hypergraph transformer to effectively integrate spatio-temporal data. 
MTNet~\cite{huang2024learning} introduces a novel "Mobility Tree" structure to capture users' check-in patterns across multiple time slots, enabling personalized next POI recommendations by learning specialized behavioral preferences for different temporal periods.

For the graph learning module, the construction of a graph is of great importance to the determination of what additional information to provide to downstream tasks. A user-POI graph, for example, can reveal users' historical activities, while a user-user graph usually contains the social network in a group. 
In our TrajGEOS model, we construct a global trajectory graph based on all users' historical check-in data and add contextual information as node features. By using hierarchical graph convolution, our graph learning module can provide additional location features and user preferences for downstream predictors.

\section{Problem Formulation} \label{sec:pf}

Let $U = \{u_1, u_2, \dots, u_N \}$ be a set of users, $L = \{l_1, l_2, \dots, l_M \}$ be a set of locations and $N$, $M$  are the total number of users and locations in a given dataset, respectively. Each location $l_i\in L$ is associated with a tuple $(c_i, lat_i, lon_i)$ that contains location category (e.g., shopping mall and restaurant), latitude and longitude. Based on these basic concepts, we hereafter introduce several key definitions used in this paper.

\textbf{Definition 3.1 (Record). } Record $r_i^k$ is a 2-tuple $(l_i^k, t_i^k)$, representing user ($u_i$)'s visited location $l_i^k$ at time $t_i^k$, where $u_i\in U$ and $l_i^k\in L$ .

\textbf{Definition 3.2 (Individual trajectory).} The trajectory of user $u_i$ is a sequence $R_i = [r_i^1, r_i^2, \dots, r_i^T ]$ that contains all historical check-in records. Due to the sparsity of users' check-in data, the record timestamp in trajectory records is uneven and there is a large time gap in most trajectory records.

In data preprocessing, we split the trajectory $R_i$ of every user $u_i$ into a set of sub-trajectories, that is, $R_i = S_i^1\oplus S_i^2 \oplus \cdots \oplus S_i^{SN_i}$ where $\oplus$ denotes concatenation, $S_i^k$ is $u_i$'s $k$-th sub-trajectory, $SN_i$ is the number of sum sub-trajectories for user $u_i$. 
The length of sub-trajectory may vary and each one contains the user's check-in within a time window. 
In this paper, we segment user's trajectory using a weekly time window as people's behavior may follow a weekly periodicity.
We then split data at the user level by assigning the first 80\% sub-trajectories of each user into the training set and the remaining sub-trajectories into the test set, denoted as $R_i^{train}$ and $R_i^{test}$, respectively.

\textbf{Definition 3.3 (Global trajectory graph).} Global trajectory graph $G$ with $M$ nodes is a directed graph constructed of all users' trajectories $\{R_1^{train}, R_2^{train}, \dots, R_N^{train}\}$. 
$V=\{v_1, v_2,\dots,v_M\}$ is node set with size $|V|$ equals to the size of location set $|L|$. And $\forall v_i \in V $ in the trajectory graph represents an actual location $l_i \in L$. 
In the edge set $E=\{e_{i \rightarrow j}, e_{i \rightarrow j}, \dots\}$, $e_{i \rightarrow j}$ represents a directed edge from $v_i$ to $v_j$, indicating the transition from location $l_i$ to location $l_j$. $e_{i \rightarrow j}$ is associated with the distance between these two locations, the sum of transition numbers, and the corresponding 24-hour flow data calculated from training data. 

The goal of the next location prediction is to predict where user $u_i$ is most like to visit next, by learning from his historical trajectories. Based on the above definitions, this task can be formally described as predicting user $u_i$'s next location $l_i^{T+1}$ based on the current trajectory $S_i^p=[ r_i^t, r_i^{t+1},\dots r_i^T ]$ and the recent trajectory $\{S_i^{p-\kappa} \}, {\kappa\in [1,\dots,p-1]}$.  
% In this paper, we set $\kappa=2$, that is, we consider check-in records within the first 1-2 weeks of the current trajectory as the recent trajectory.
In this paper, we set $\kappa=2$, that is, we consider check-in records within the preceding 2\textasciitilde 3 weeks as the recent trajectory.

\section{METHODOLOGY}

\subsection{Model Structure Overview}
\begin{figure*}[htbp]
  \centering
  \includegraphics[width=0.90\linewidth]{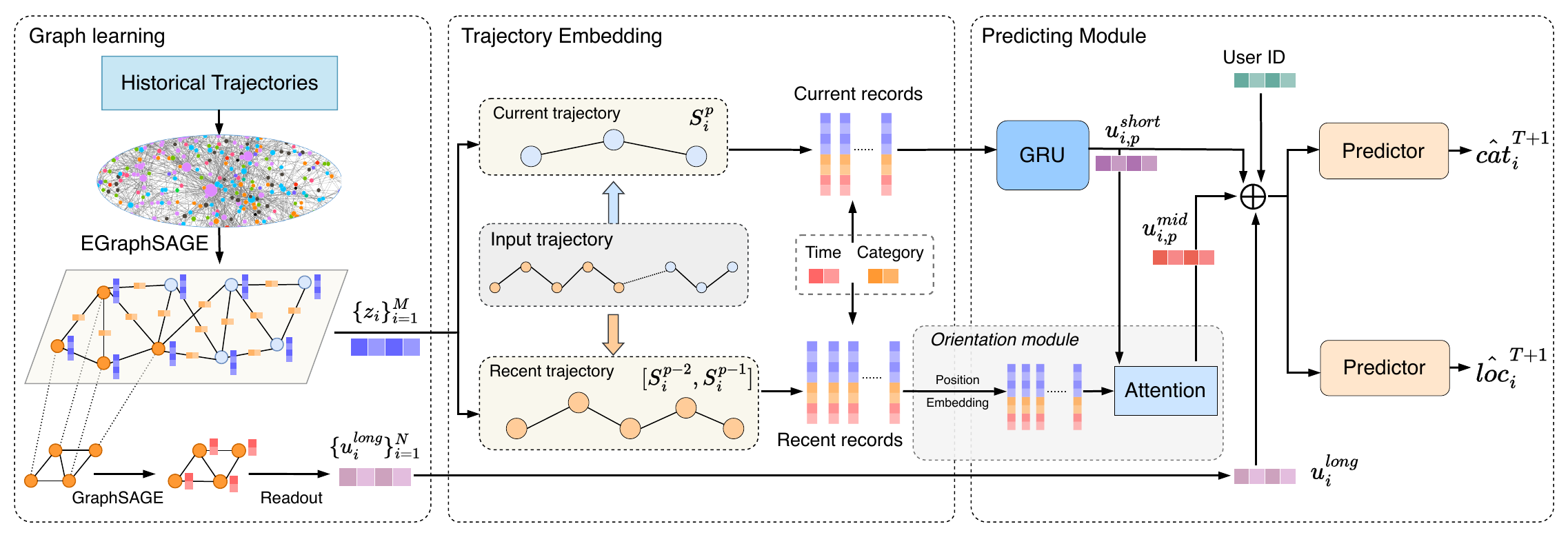}
  \caption{The architecture of TrajGEOS. It contains three modules: graph learning module, trajectory embedding module, and prediction module. }
  % The graph learning module performs graph convolution on trajectory graphs to obtain location embedding and the user's long-term preference. The trajectory embedding module embeds trajectories with additional check-in data such as location category and visiting time. In the prediction module, we employ GRU and an attention-based orientation module to capture the user's short- and mid-term preferences and apply a multi-task strategy to predict both the next category and the next location.} 
  \label{fig:model_architecture} 
\end{figure*}

Figure \ref{fig:model_architecture} illustrates the framework of our TrajGEOS model, which consists of three key components. Firstly, we construct a trajectory graph based on the check-in records of all users in the training data and then apply hierarchical graph convolution operation to capture the embeddings of each location and each user, enriched with additional spatial and contextual information. 
% Second, within the context embedding module, we utilize the embedding layer to encode user id and contextual information. 
Second, within the trajectory embedding module, we utilize the embedding layer to encode the contextual information of the user's historical check-in locations. 
By incorporating additional embeddings from our graph learning module, we can obtain encoded trajectory sequences. 
Thirdly, in the prediction module, we design a multi-task predictor that employs a shared GRU to capture patterns in sub-trajectories and utilizes two independent MLPs to predict the next location and category. 
In the following subsections, we will provide further elaboration on the TrajGEOS model. 

\subsection{Learning with Trajectory Graph}  

\textbf{Initialization in trajectory graph.} Given the trajectory graph, we first initialize its node features and edge features. 
The raw features of a location $l_i$ include its identity, category $c_i$ and coordinate $(lat_i,lon_i)$. 
That is, the initial node feature of $l_i$ is: 
\begin{equation}
h_i^0 = Emb(l_i) \;||\;Emb(c_i) \;||\; lat_i \;||\; lon_i \label{eq:init_h0}
\end{equation}
where $Emb(l_i)$, $Emb(c_i)$ represent the embedding of location ID and the embedding of category. 
And $Emb(\cdot)$ represents the basic Embedding layer.

As for the initialization of edges, for every edge $e_{i\rightarrow j}$, 
we calculate the distance $distance_{i\rightarrow j}$ between $l_i$ and $l_j$ and 
count the sum transition numbers $trans_{i\rightarrow j}$ from $l_i$ to $l_j$ from all training data. 
Moreover, for every $e_{i\rightarrow j}$, we formulate 
a 24-dim flow vector $flow_i = [n_0, n_1, \cdots n_{23}]$. 
The $k$-th dim in the $flow$ of $e_{i\rightarrow j}$ represents the record number of transitions from $l_i$ to $l_j$ at hour $k$ in all training data.
With these features, we get the initial embedding of edge $e_{i\rightarrow j}$ like the following equation shows:
\begin{equation}
\mathcal{E}_{i\rightarrow j}^0 = trans_{i\rightarrow j} \;||\; distance_{i\rightarrow j} \;||\;  flow_{i\rightarrow j} \label{eq:init_edge}
\end{equation}

\textbf{Graph convolution on global trajectory graph.}
To make full use of node features and edge features in the global trajectory graph, we adopt GRAPE~\cite{you2020handling} (hereafter referred to as EGraphSAGE) as the lower graph convolution structure to learn location embeddings. 
EGraphSAGE not only leverages the features of neighboring nodes and edges in the process of message passing but also simultaneously updates node and edge embeddings. 
In global trajectory, the initial node feature includes identity, coordinates, and category (for some datasets) and the initial edge feature includes distance and flow data between locations. So we not only update the node features but also update the edge feature using the updated node features in each EGraphSAGE layer. 
The operation of the EGraphSAGE layer is like the following equation shows: 
\begin{align}
h_{\mathcal{N}(s, \epsilon)}^k & = {\rm {MEAN}} \left(\sigma \left( \mathbf{W}_1^{k-1} \cdot {\rm {CONCAT} } (h_t^{k-1}, \mathcal{E}_{t \rightarrow s}^{k-1}) \;|\; \forall t \in \mathcal{N} (s, \epsilon) \right) \right) \label{eq:global_node_update1} \\
h_s^{k} & = \sigma \left( {\mathbf W}_2^k \cdot {\rm CONCAT}(h_s^{k-1}, h_{\mathcal{N}(s, \epsilon)}^k) \right) \label{eq:global_node_update2} \\
\mathcal{E}_{t \rightarrow s}^{k} & = \sigma \left(\mathbf{W}_3^k \cdot {\rm CONCAT} (\mathcal{E}_{t \rightarrow s}^{k-1} , h_{t}^k, h_{s}^k) \right)  \label{eq:global_edge_update} 
\end{align}

where $\mathbf{W}_1^k, \mathbf{W}_2^k, \mathbf{W}_3^k$ are learnable parameters, 
$\epsilon$ is the edge dropout ratio in the global trajectory graph. 
The global trajectory graph is a complex and humongous structure that records all users' mobility traces in a specific region. Transition backbones like core metro or bus stations usually carry a large flow of human mobility.
As a result, their corresponding nodes in the global trajectory graph always have large degrees.
Operating graph convolution on these nodes makes these central nodes update their embedding based on numerous neighborhoods. As a result, makes it easier to over-smoothing.
So we set edge dropout ratio $\epsilon$ to $0.5$, which is an experiential value used in dropout layers, in our experiments to avoid complex network structure and alleviate the over-smoothing problem. 
$s, t$ represent nodes $v_s, v_t$ in graph, $\mathcal{N}(s, \epsilon)$ are the neighbor set of node $v_s$ with edge dropout ratio $\epsilon$, 
$h_s^k$ is node $v_s$'s embedding in the $k$-th layer. 

We next extract $h_i^2 \in \mathbb{R}^{d_{Gl}}$ as the output of the 2-layer graph convolution on the global trajectory graph, where $d_{Gl}$ is the dimension of node embedding in the graph learning module. 
Then we concatenate $h_i^2$ with the initial node feature $h_i^0$ and feed the concatenated vector into a dense layer utilized in~\cite{yang2022getnext}. The output can be denoted as:
\begin{equation}
    z_i = \sigma(\mathbf{W}_4 [h_i^2 \,||\, h_i^0] + b ) \label{eq:fusion}
\end{equation}
where $\mathbf{W}_4$ and $b$ are trainable parameters. We regard $z_i \in \mathbb{R}^{d_{Gl}}$ as the final embedding of location $l_i$. 
Now that after graph learning, we can collect the embeddings of all locations, and then have the embedding matrix $\mathbf{Z}\in \mathbb{R}^{M\times d_{Gl}}$ as the location feature for downstream modeling. 

\begin{figure}[htbp]
  \centering
  \includegraphics[width=0.80\linewidth]{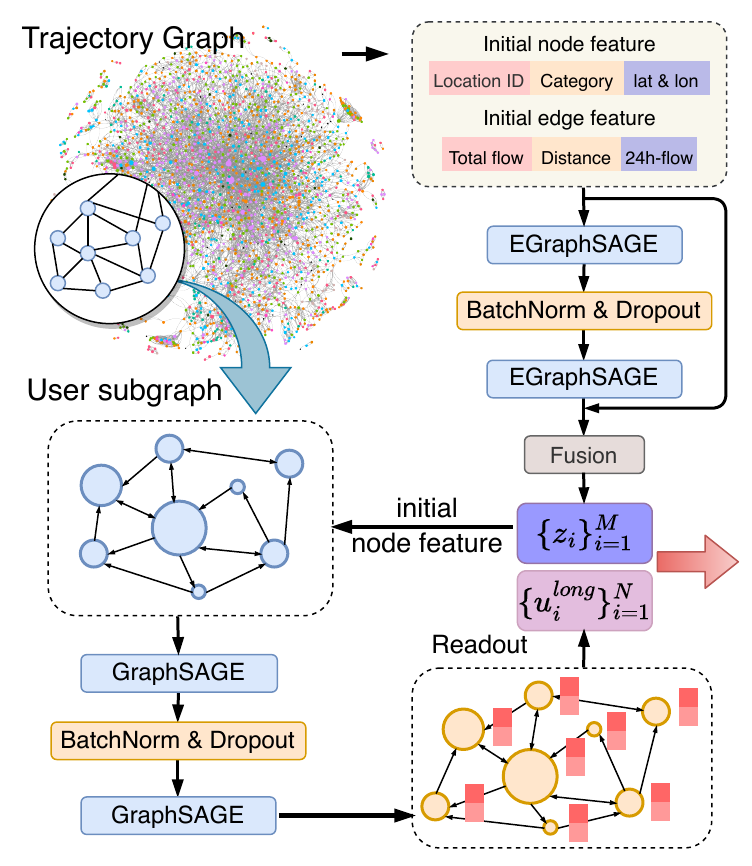}
  \caption{Hierarchical graph convolution in the graph modeling module.}
  \label{fig:graph-learning} 
\end{figure}

\textbf{Graph learning on user's trajectory graph.} To capture the user's historical mobility patterns, we further apply graph convolution on user subgraphs and calculate the subgraphs' readout as the user's stable long-term preference. 
More specifically, the calculation consists of the following steps. 
Firstly, for a specific user, say $u_i$, we extract his subgraph $G_i$ from the trajectory graph based on his historical check-in records $R_i^{train}$. 
In $G_i$, the node set is $V_i$, corresponding to $u_i$'s visited location set $L_i$, the edge set is $E_i$. 
The initial node embedding of $v_{j}\in V_i$ is the $z_{j}$ obtained from the global location embedding matrix $\mathbf{Z}$.
Next, we use 2-layer GraphSAGE~\cite{hamilton2017inductive} to update node embeddings in the user's trajectory graph and finally get $\{h_{i,j}^2\}$ where ${l_j\in V_i}$ as location embeddings unique to user $u_i$. 
For user $u_i$, the updating of node embeddings in the GraphSAGE layer is like the following shows: 
\begin{align}
h_{i, \mathcal{N}(s, \epsilon)}^k & = {\rm{mean}} \left(\{h_{t}^{k-1}, \forall t \in \mathcal{N}(s, \epsilon)\} \right) \label{eq:subgraph_conv1} \\
% h_{i,s}^{k} & = \sigma \left( {\mathbf W}_5^k \cdot {\rm CONCAT}(h_s^{k-1}, h_{\mathcal{N}(s, \epsilon)}^k) \right) \label{eq:subgraph_conv2}
h_{i,s}^{k} & = \sigma \left( {\mathbf W}_5^k \cdot {\rm CONCAT}(h_{i-1, s}^{k-1} \,,\, h_{i, \mathcal{N}(s, \epsilon)}^k) \right) \label{eq:subgraph_conv2}
\end{align} 
where ${\mathbf W}_5^k$ is a trainable parameter, while $s, t, \epsilon$ have the same meaning as in the equations of EGraphSAGE layers. 
$h_{i,s}^{k}$ represents the embedding of node $v_s$, which is unique to user $u_i$, in the $k$-th layer.
It is worth noting that the graph convolutions on user subgraphs do not affect the node embeddings in the global trajectory graph. 

Then, we calculate the visiting weight for every location within the user's trajectory subgraph. Assume that user $u_i$'s visiting set is $V_i$ and $\eta_{i,j}$ is the number of records user $u_i$ visited location $l_j$. The sum of records of user $u_i$ is $RN_i = \sum_{l_j \in V_i}  \eta_{i,j} $. 
For user $u_i$, the visiting weight of the location $l_j$ is $w_{i,j} = {\eta_{i,j}} / { RN_i }$.

The visiting weight-based readout of user $u_i$'s trajectory subgraph is calculated with the following formula:
\begin{equation}
\label{eq:user_readout}
u_i^{long} = \sum_{l_j\in V_i} w_{i,j} \cdot h_{i, j}^2
\end{equation}
where $h_{i, j}^2 $ is the embedding of location $l_j$ unique to user $u_i$.  
We regard the user embedding $u_i^{long}$, which is obtained from the user's trajectory graph, as user $u_i$'s long-term preference, and utilize it in the downstream prediction task. 
The detailed structure of the graph learning module is shown in Figure~\ref{fig:graph-learning}, and the whole process of the graph learning task is formally shown in Algorithm~\ref{alg:graph_learning}.

\subsection{Embedding and capturing of trajectory}
We encode the raw trajectory records into embedding sequences and employ sequential models to capture transition patterns. This subsection presents the trajectory captures for the downstream sequential model. 

\textbf{Embedding layers.} Every data record $r_i^k$ for user $u_i$ is a 2-tuple $(l_i^k, t_i^k)$, including location id $l_i^k$ , visiting time $t_i^k$, and the location category $c_i^k$ associated with $l_i^k$. 
We use basic embedding layers to encode user ID and category ID information. These id-based embeddings play an important role in distinguishing different check-in records. 
The visiting time $t_i^k$ includes both visiting-weekday and visiting-hour information, and some check-in records might be periodic in time. Thus, we use Time2Vec~\cite{kazemi2019time2vec} to model the periodicity and get $Time2Vec(t_i)$ for every timestamp. 
Additionally, $Emb(u_i)$, $Emb(c_i)$, $Time2Vec(t_i)$ are used to represent the embedding of user $u_i$, category $c_i$, and time $t_i$. 

\textbf{Trajectory Embedding.} 
As we have introduced in Section~\ref{sec:pf}, we leverage the user's historical trajectory from the past three weeks to predict their next visiting location instead of relying on their entire historical trajectory. 
Here we do not make a strict distinction but uniformly use 'trajectory' to refer to the historical trajectory data used in prediction.
As previously mentioned, every user's trajectory is a sequence of check-in records that include user id, location id, category id, and visiting time. 
Assume user $u_i$ has one record $r_i^k$ that visited location $l_j$ at time $t_j$, where $c_j$ is the category of location $l_j$. 
Then we can embed this single record based on the embeddings presented above: 
\begin{equation}
    Q(r_i^k) = z_j \,||\, Emb(c_j) \,||\, Time2Vec(t_j)
\end{equation}
Where $z_j$, which represents the location identity, is the output of lower graph convolution module, $Emb(c_j)$ and $Time2Vec(t_j)$ are the embedding of location category and visiting time. So the encoded trajectory $Q(S_i^p)$ is:
\begin{equation}
    Q(S_i^p) = [ Q(r_i^{k}), Q(r_i^{k+1}), \dots ]
\end{equation}
Where the actual trajectory is $S_i^p=[r_i^k, r_i^{k+1}, \dots ]$ .

To better model the travel patterns of users across different historical periods, we partition every user's historical trajectory into two segments. 
Specifically, we define the check-in records within the latest week as the current trajectory and those within the preceding 2\textasciitilde 3 weeks as the recent trajectory.

We utilize GRU to directly model the user's encoded current trajectory and interpret its output as the user's short-term preference for the next movement. 
In downstream prediction, the output of GRU not only directly contributes to the prediction, but also plays a role in calculating the user's mid-term preference as a query, which will be further elaborated in the next subsection.

\subsection{Predicting Module}

\subsubsection{Orientation Module} 
Although long- and short-term preferences are commonly utilized in many existing methods \cite{wu2019long,feng2018deepmove,sun2020go}, we propose that incorporating the user's recent historical trajectory is crucial for complementing their travel preference information. 
Therefore, we design an orientation module that leverages encoded recent historical trajectory sequences and the outputs of GRU to capture the user's mid-term preferences. 

\textbf{Recent historical trajectory.} 
Assume the target next location of user $u_i$ is $l_i^{T+1}$ and $u_i$'s latest check-in record is located in sub-trajectory $S_i^p=[ r_i^k, r_i^{k+1},\dots r_i^T ]$, which is regarded as the current trajectory. 
We select the recent sub-trajectory $[S_i^{p-2}, S_i^{p-1}]$ as the recent historical trajectory for $u_i$ to predict the next location $l_i^{T+1}$. 
Because we split trajectory by week in this paper, the recent historical sequence of user $u_i$ actually corresponds to his historical records in the preceding 2\textasciitilde 3 weeks. 

\textbf{Position Embedding.}
We encode the user's recent historical trajectory in the same format as input sequences of GRU and get corresponding encoded sequences. 
Then we add positional embeddings to these encoded recent historical trajectories. We use sine and cosine functions proposed in transformer~\cite{vaswani2017attention}:
% remove detailed introduce to position embedding method
% \begin{align}
% PE(pos, 2i) & = \sin(pos/10000^{2i/d_{model}}) \\
% PE(pos, 2i+1) & = \cos(pos/10000^{2i/d_{model}})
% \end{align}
% Where $pos$ is the position of location in trajectory, $2i$ and $2i+1$ represent the particular dimension on the location embedding and $d_{model}$ is the size of the location embedding dimension. 
With the incorporation of additional position embedding, the following module can grasp the relative positions in encoded recent historical trajectories. 

Subsequently, we leverage the attention mechanism to compute the user's mid-term preference based on the enforced encoded trajectories with position embedding and GRU module outputs. 
Assume $u_{i, p}^{short}$ is the output of GRU when the input is $S_i^p$, and the user $u_i$'s encoded recent trajectory is 
$\{Q(r_i^j)\}$, where $r_i^j \in [S_i^{p-2}, S_i^{p-1}]$.
The output of orientation module $u_{i,p}^{mid}$ is defined as:
\begin{align}
% \alpha_j & = MLP( Q(r_i^j) \,||\, u_{i, p}^{short} ) \\
% \beta_j & = \frac{exp(\alpha_j)}{\sum_{l=1}^{k} exp(\alpha_l)} \\
% u_{i,p}^{mid} & = \sum_{j=1}^{k} \beta_j \cdot Q(r_i^j) 
\beta_j & = MLP\left( Q(r_i^j) \,||\, u_{i, p}^{short} \right) \\
u_{i,p}^{mid} & = \sum_{j=1}^{k} \frac{exp(\beta_j)}{\sum_{l=1}^{k} exp(\beta_j)} \cdot Q(r_i^j) 
\end{align}
We regard $u_{i,p}^{mid}$ as the user $u_i$'s mid-term preference when his current trajectory is $S_i^p$ and use it for the next location prediction task.

\subsubsection{Multi-task learning in TrajGEOS} 
We design a multi-task learning strategy for TrajGEOS like Figure \ref{fig:model_architecture} shows. 
The next category prediction task is regarded as an auxiliary task, which shares the same GRU outputs with next location prediction task and utilizes another MLP to predict the next category.

\textbf{Predicting next category and location} For user $u_i$, set $Emb(u_i)$ as the encoded user id generated by basic embedding layer.
Assume $u_i$'s current trajectory is $S_i^p$.
Then user $u_i$'s preference $\phi_i^T$ used for the prediction of next category $cat_i^{T+1}$ and $loc_i^{T+1}$ can be represented as:
\begin{equation}
    \phi_i^T = u_{i, p}^{short} \;||\; u_{i,p}^{mid} \;||\; u_i^{long} \;||\; Emb(u_i)
\end{equation}
Where $u_{i, p}^{short}$ is the output of GRU that processed $u_i$'s encoded current trajectory $S_i^p$, $u_{i,p}^{mid}$ is the corresponding mid-term preference generated by orientation module and $u_i^{long}$ is user's long-term preference from graph learning module.

We consider the next place and category prediction job as a multi-classification task, that can be formally formulated as:
\begin{align}
    % \bm{cat}_i^{T+1} & = MLP_c(\phi_i^T) \\
    % \widehat{cat}_i^{T+1} & = {\rm argmax} \left( \bm{cat}_i^{T+1} \right) \\
    % \bm{loc}_i^{T+1} & = MLP_l(\phi_i^T) \\
    % \widehat{loc}_i^{T+1} & = {\rm argmax} \left( \bm{loc}_i^{T+1} \right)
    \widehat{cat}_i^{T+1} & = {\rm argmax} \left( MLP_c(\phi_i^T) \right) \\
    \widehat{loc}_i^{T+1} & = {\rm argmax} \left( MLP_l(\phi_i^T) \right)
\end{align}
% Where the dimension of $\bm{loc}_i^{T+1}$ is the size of location set and the dimension of $\bm{cat}_i^{T+1}$ is the size of category set.

\textbf{Loss function.} As a multi-task learning model, our loss function includes the cross entropy losses of both the next category prediction $\mathcal{L}_c$ and the next location prediction $\mathcal{L}_l$. For example, the cross entropy loss of next location prediction $\mathcal{L}_l$ is:
\begin{equation}
    \mathcal{L}_l = -\sum_{j=1}^M y_j \log (p_j)
\end{equation}
Where $M$ is the dimension of $\bm{loc}_i^{T+1}$, $y_j$ equals to 1 only when $loc^{T+1}==l_j$ and $p_j$ is the $j$-th dimension of $softmax( \bm{loc}_i^{T+1} )$.
The total loss of TrajGEOS is:
\begin{equation}
    \mathcal{L} = \alpha\mathcal{L}_l + (1-\alpha) \mathcal{L}_c
\end{equation}
Where $\alpha$ is the hyperparameter that controls the weights of different tasks. 
In this paper, we set $\alpha$ to $0.7$ to obtain the optimal $Recall@1$ for the next location prediction task. 
Further details about the experiments of $\alpha$ can be found in Table~\ref{tab:alpha_parameter}.

\section{Experiments}

\subsection{Experimental Setup}

\subsubsection{Datasets.} 

We conduct experiments on three public check-in datasets:  NYC~\cite{yang2014modeling}, TKY~\cite{yang2014modeling}, and Gowalla~\cite{cho2011friendship} in Dallas. The NYC dataset was collected from April 2012 to February 2013 in New York City, and TKY was from Tokyo during the same period. Data in Dallas consists of the public check-in data with time and location information, collected from Feb. 2009 to Oct. 2010. 
Records in NYC and TKY contain fields including user ID, location ID, location category ID, GPS coordinate, and timestamp. The record fields of Dallas are similar to those in NYC except that there is no category data. So we do not use category information to encode trajectories in the Dallas dataset. 
For all three datasets, we exclude unpopular locations and the outlier users with less than 10 records, in line with~\cite{sun2020go, feng2018deepmove}. Also, we merge the contiguous records with the same user and location at the same hour. After the preprocessing, we further process the data for our TrajGEOS model by splitting the user's entire trajectory into sub-trajectories according to week. 
Every sub-trajectory should contain check-in records of a single user for at least two instances in a week. Additionally, every user must have at least five sub-trajectories following the setting of~\cite{feng2018deepmove}.

The baseline models have specific data input requirements, thus we utilize the source code provided by the authors for data processing. Overall, the datasets exhibit minimal differences, allowing for a meaningful comparison.
Additionally, GETNext~\cite{yang2022getnext} and MTNet~\cite{huang2024learning} cannot be applied to the Dallas dataset as it requires category information in training. The statistical information of data used by different models is shown in Table~\ref{tab:data_info}.

% Dataset information table
\begin{table}[!htbp]
\centering
\caption{Dataset information}
\label{tab:data_info}
\resizebox{\linewidth}{!}{
\begin{tabular}{cccccccccc} 
\hline
          & \multicolumn{3}{c}{NYC}    & \multicolumn{3}{c}{TKY}    & \multicolumn{3}{c}{Dallas}  \\ 
\cline{2-10}
          & user  & location & records & user  & location & records & user  & location & records          \\ 
\hline
RAW       & 1,083 & 38,333   & 227,428 & 2,293 & 61,858   & 573,703 & 5,894 & 5,767    & 167,016          \\
Processed & 1,083 & 4,638    & 139,183 & 2,293 & 7,222    & 427,746 & 2,412 & 5,642    & 146,117          \\ 
\hline
FPMC-D    & 1,083 & 4,638    & 138,099 & 2,293 & 7,222    & 425,450 & 2,412 & 5,642    & 143,704          \\
FPMC-W    & 1,083 & 4,638    & 138,098 & 2,293 & 7,222    & 425,445 & 2,412 & 5,642    & 143,704          \\
DeepMove  & 1,061 & 4,627    & 111,968 & 2,284 & 7,206    & 333,215 & 1,193 & 5,346    & 93,911           \\
GeoSAN    & 1,073 & 4,611    & 138,229 & 2,289 & 7,209    & 427,157 & 2,300 & 5,357    & 142,980          \\
% Flashback & 439   & 4,093    & 96,771  & 1,451 & 6,998    & 366,604 & 340   & 5,434    & 82,287           \\
LSTPM     & 1,019 & 4,614    & 121,148 & 2,233 & 7,201    & 395,192 & 954   & 5,366    & 103,664          \\
GETNext   & 1,066 & 4,621    & 131,920 & 2,280 & 7,200    & 414,993 & -     & -        & -                \\
MTNet     & 1,066 & 4,621    & 131,920 & 2,280 & 7,200    & 414,993 & -     & -        & -                \\
TrajGEOS  & 1,065 & 4,635    & 131,874 & 2,280 & 7,204    & 414,855 & 1,357 & 5,428    & 118,069          \\
\hline
\end{tabular}
}
\end{table}

% Our model shares the same preprocessed data with the baselines, yet, due to the differences in the modeling methods, the actual data used by the models might differ from one another. 
% The statistical information of data used by different models is shown in Table~\ref{tab:data_info}. 
% It is noteworthy that Flashback~\cite{yang2020location} uses a stricter filtering policy and limits the minimal records for users to 100, resulting in a smaller data size. 
% Additionally, GETNext~\cite{yang2022getnext} and MTNet~\cite{huang2024learning} cannot be applied to the Dallas dataset as it requires category information in training.

%  Experiment results
\begin{table*}[!htbp]
\centering
\caption{Results of next location prediction task}
\label{tab:main_result}
\resizebox{\linewidth}{!}{
\begin{threeparttable}
\begin{tabular}{ccccc|cccc|cccc}
\hline
\multirow{2}{*}{Model} & \multicolumn{4}{c|}{NYC}    & \multicolumn{4}{c|}{TKY}      & \multicolumn{4}{c}{Dallas}     \\ 
\cline{2-13}
                 & R@1        & R@5        & R@10       & M@10          & R@1        & R@5        & Recall10        & M@10          & R@1        & R@5        & R@10       & M@10           \\ 
\hline
FPMC-D                 & 0.1598          & 0.4581          & 0.5816          & 0.2842          & 0.1279          & 0.3575          & 0.4701          & 0.2243          & 0.0596          & 0.1653          & 0.2300          & 0.1052           \\
FPMC-W                 & 0.1469          & 0.4437          & 0.5788          & 0.2714          & 0.1225          & 0.3425          & 0.4565          & 0.2158          & 0.0550          & 0.1626          & 0.2275          & 0.1020           \\
% [WWW, 2018]
DeepMove    & 0.2007          & 0.4108          & 0.4685          & 0.2890          & 0.1605          & 0.3244          & 0.3924          & 0.2305          & 0.0905          & 0.1884          & 0.2294          & 0.1320           \\
% [KDD, 2019]
GeoSAN        & 0.1445          & 0.3886          & 0.5592          & 0.2537          & 0.2049          & 0.4557          & \underline{0.5968}          & 0.3155          & 0.0670          & 0.2065          & 0.2935          & 0.1255           \\
% [IJCAI, 2020]
% Flashback  & 0.2230       & 0.5250        & 0.6409        & 0.3494        & 0.2071        & \underline{0.4866}        & 0.5828         & 0.3235        & 0.0728        & 0.1137        & 0.1396        & 0.0906  \\
% [AAAI. 2020]
LSTPM      & 0.2431        & \underline{0.5478}        & \underline{0.6645}        & 0.3714        & 0.2103        & 0.4602        & 0.5588         & 0.3155        & 0.1313        & \underline{0.2567}        & \underline{0.3269}        & \underline{0.1842}  \\
% [SIGIR, 2022]
GETNext    & 0.2406          & 0.5323          & 0.6220          & 0.3625          & 0.2180         & 0.4583          & 0.5550          & 0.3203          & -               & -               & -               & -                \\
MTNet      & \underline{0.2635} &  0.5445 & 0.6230 & \underline{0.3817} & \underline{0.2388} & 0.4816 & 0.5698 & \underline{0.3414}  & - & - & - & - \\ 
TrajGEOS        & \textbf{0.2721} & \textbf{0.5740}   & \textbf{0.6716}   & \textbf{0.3982}  & \textbf{0.2490} & \textbf{0.5043}   & \textbf{0.5997}   & \textbf{0.3564}    & \textbf{0.1326} & \textbf{0.2646}    & \textbf{0.3280}     & \textbf{0.1893}           \\ 
\hline
TrajGEOS-woGraph       & 0.2622          & 0.5678          & 0.6651          & 0.3899          & 0.2017          & 0.4531          & 0.5484          & 0.3075          & 0.1071          & 0.2423          & 0.2960          & 0.1645           \\
TrajGEOS-onlyGraph     & 0.2653          & 0.5540          & 0.6444          & 0.3859          & 0.2439          & 0.4936          & 0.5864          & 0.3491          & 0.1227          & 0.2411          & 0.2978          & 0.1737           \\
TrajGEOS-woShort       & 0.2318          & 0.5434          & 0.6566          & 0.3615          & 0.2006          & 0.4666          & 0.5712          & 0.3126          & 0.1093          & 0.2462          & 0.3063          & 0.1680           \\
TrajGEOS-woMid         & 0.2638          & 0.5693          & 0.6692          & 0.3910          & 0.2478          & 0.5054          & 0.6046          & 0.3565          & 0.1310          & 0.2665          & 0.3290          & 0.1887           \\
\hline
\end{tabular}
\end{threeparttable}
}
\end{table*}

\subsubsection{Baselines.} 
In recent years, some effective solutions have been proposed to tackle the next location prediction problem. Here we select the following baselines for comparison:
\begin{itemize}
    \item FPMC~\cite{rendle2010factorizing} is a Markov-based model utilizing matrix factorization to process users' personalized transition matrix and learn individual transition patterns.
    \item DeepMove~\cite{feng2018deepmove} is an attentional recurrent model using attention mechanism and GRU to capture long- and short-term transition patterns.
    % \item Flashback~\cite{yang2020location} designs a general RNN architecture by considering flashbacks on RNN's hidden states to deal with sparse user traces.
    \item LSTPM~\cite{sun2020go} uses the non-local network to model the user's long-term preference, and leverages geo-dilated RNN to capture geographical relations among non-consecutive locations.
    \item GeoSAN~\cite{lian2020geography} is a seq2seq model that uses attention-based networks as the encoder and decoder, and designs geography-aware negative samplers to use spatial information.
    \item GETNext~\cite{yang2022getnext} applies graph learning on trajectory graph to capture the global transition patterns among locations and leverages a transformer architecture to facilitate the prediction of users' next movements.
    \item MTNet~\cite{huang2024learning} introduces a novel tree structure to hierarchically describe the users' preferences across varying temporal periods.
\end{itemize}

\subsubsection{Evaluation metrics.} We select the two most commonly used metrics in next location prediction tasks, recall and mean reciprocal rank to evaluate models' performances. 
% As previously introduced, our objective is to predict the next location and category of each sample based on its historical data. 
Given a test data with $m$ samples, $Recall@K$ and $MRR@K$ are defined as follows:
\begin{align}
R@K & = \frac{1}{m}\sum_{i=1}^m \mathbbm{1}(rank\leq K) \\
M@K & = \frac{1}{m} \sum_{i=1}^m \frac{\mathbbm{1}(rank\leq K)}{rank}
\end{align}
where $m$ is the number of test data, $\mathbbm{1}$ is an indicator function and returns 1 if the condition is true, otherwise 0. $rank$ is the index of the true predicted location or category in the recommended order list. 
We finally set $K=1, 5, 10$ for the Recall metric and $K=10$ for MRR. For both R@K and M@K, a larger value means better performance.

% move detailed experimental setting to appendix 

\subsection{Main Results}
Our TrajGEOS model employs a multi-task learning approach, with the primary task being next location prediction and the secondary task being next category prediction. Most baseline models except GETNext and MTNet, however, only focus on predicting the next location without considering the category. Therefore, we first compare the performance of the next location prediction task in Table~\ref{tab:main_result}. 
Furthermore, since the baseline models do not directly predict the category of user's next location, we utilize a mapping between location and category that is derived from raw check-in data to convert their predicted location IDs into corresponding categories. We then evaluate the performance of next category prediction and present the results in Table~\ref{tab:cat_result}. 
% Last but not least, it is noteworthy that 
The results of our models are computed through averaging across five independent runs, ensuring the stability of TrajGEOS experimental results. And we report $Recall@1$ ($R@1$), $Recall@5$ ($R@5$), $Recall@10$ ($R@10$), and $MRR@10$ ($M@10$) for all datasets.

% \subsubsection{Next location prediction} 

For all datasets, our model outperforms the baseline models in the next location prediction task. In terms of $Recall@1$ in the location prediction task, TrajGEOS achieves 27.2\%, 24.9\%, and 13.2\% in NYC, TKY, and Dallas datasets, while the $Recall@1$ values for the baseline model with the best performance are 26.4\%, 23.8\%, and 13.1\%. Among the baseline methods, MTNet achieves superior performance in $Recall@1$ and $MRR@10$ but does not ensure optimal results in $Recall@5$ and $Recall@10$. In contrast, TrajGEOS consistently maintains a leading performance across all four evaluation metrics.

% todo
% As Table~\ref{tab:main_result} shows, the overall performance of location in the Dallas dataset is comparatively inferior to that of the NYC and TKY datasets. We look into the distribution of coordinates in these 3 datasets and find that the locations in Dallas are widely spread in a square area over 15,000 $km^2$ while locations in NYC and TKY dataset are squeezed in a square area of about 2,000 $km^2$, which means the records in Dallas is faced with a more serious problem of spatial sparsity. Despite the severe geographical sparsity, TrajGEOS still outperforms all baseline models on the Dallas dataset, indicating it would perform better on sparse data as well.

% \subsubsection{Next category prediction}
% TODO move to appendix

\subsection{Ablation Study}

To examine the contributions of different components in TrajGEOS, we design four ablation models: 

\begin{itemize}
	\item TrajGEOS-woGraph: eliminates the graph learning module and disregards any long-term user preferences learned from their trajectory graph. 
	% \item TrajGEOS-onlyGraph: only uses graph learning module to learn the embedding of locations in trajectory and user's long-term preference. It does not use embedding layers to model the location category in the trajectory embedding module and does not use the embedding of raw user ID in the predicting module. 
        \item TrajGEOS-onlyGraph: only uses graph learning module to learn the embedding of locations in trajectory and user's long-term preference. It does not use embedding layers to model the location category and does not use the embedding of the raw user ID. 
	% \item TrajGEOS-woShort: does not incorporate GRU output, which models user's current trajectory, into predictors for both next category and next location. As a result, it neglects the user's short-term preferences.
        \item TrajGEOS-woShort: does not calculate the user's short-term preference for the downstream predictor.
	\item TrajGEOS-woMid: deletes the orientation module and does not calculate the user's mid-term preference for the downstream predictor. 
\end{itemize} 

Table \ref{tab:main_result} presents the results of ablation studies that focus on the location prediction task. We find that TrajGEOS-woShort has the poorest performance among the models, which means users' short-term preferences are of great importance in modeling the transition pattern. 
The TrajGEOS-onlyGraph experiment reveals that the embedding of users based on their historical trajectory graphs (user's long-term preference) cannot fully substitute for raw user ID features in distinguishing between different users. Moreover, incorporating category information from a user's historical trajectory enhances the accuracy of predicting their next location. 

% TODO 
% Comparing the performance of TrajGEOS and its ablation versions, we can confirm that our trajectory graph learning module and orientation module do provide more valuable information to predictor and contribute to the next location task.

\subsection{Visualization}

To intuitively compare the performance of different models, We visualize the prediction results of TrajGEOS along with the other five baseline models on the datasets of NYC and TKY and also check the impact of various factors on the prediction task. 
 
\begin{figure}[!htbp]
  \centering
  \includegraphics[width=0.98\linewidth]{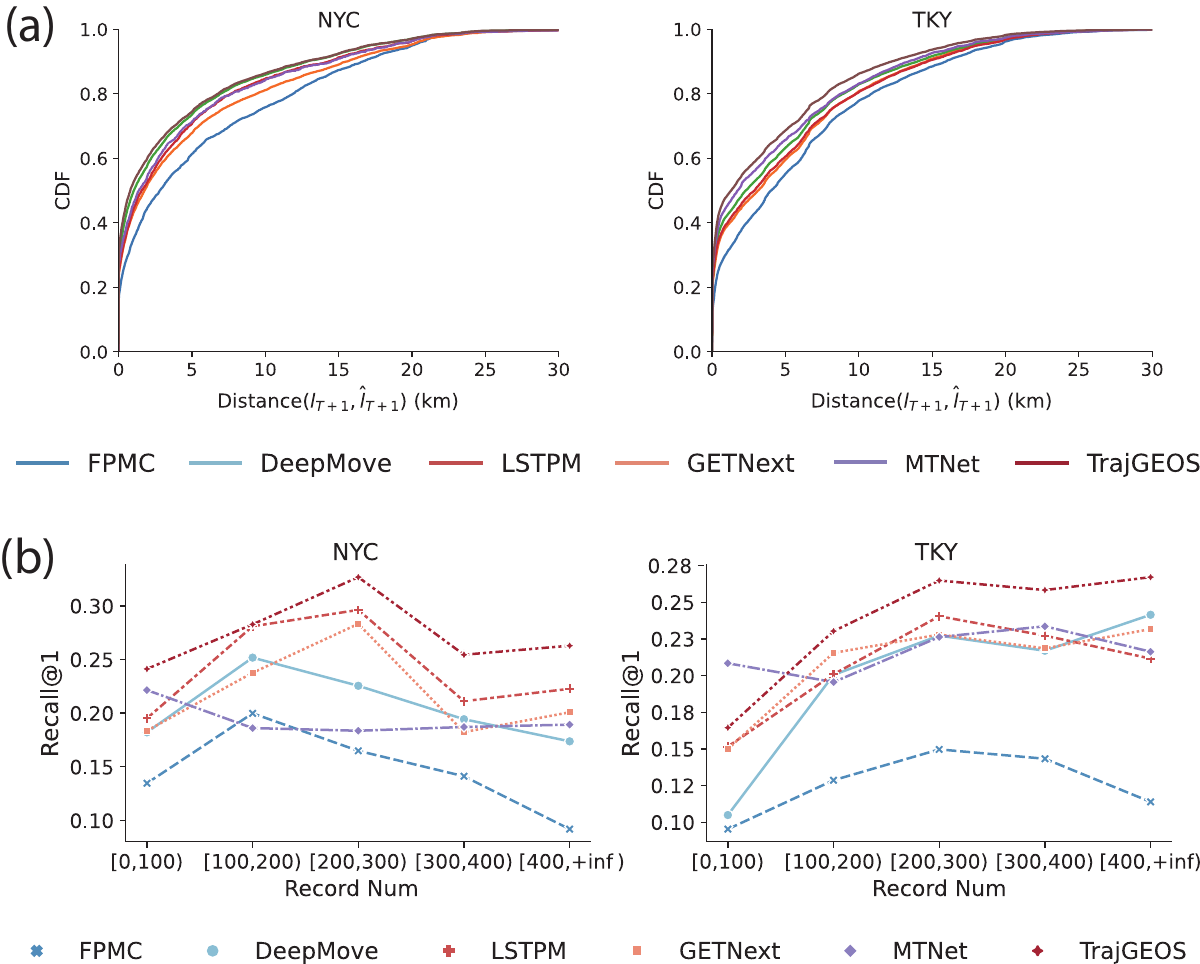}
  \caption{(a) The cumulative distribution function of distance error($\hat{l}^{T+1}, l^{T+1}$). (b) The relationship between users' record num and the predicted accuracies of their next locations.}
  \label{fig:distance_error_and_recordNum_metric} 
\end{figure}

We plot Figure~\ref{fig:distance_error_and_recordNum_metric} (a) to explore the distribution of distance errors, which are calculated based on the distance between predicted top-1 locations and target locations, generated by the prediction results of the models. 
The cumulative distribution function (CDF) curve of the distance deviation for TrajGEOS lies above that of the baseline model, indicating that the distance deviations of TrajGEOS are concentrated in smaller values. This suggests that the predictions made by TrajGEOS are spatially closer to the ground truth, thereby demonstrating superior predictive performance. 

Besides, a user's activity level greatly impacts the next location prediction task. Since the number of users' check-in records is related to their activity, we categorize users according to the number of their historical records and look into the average performance in different user groups. Results in Figure~\ref{fig:distance_error_and_recordNum_metric} (b) show that more historical records are beneficial to the modeling of a user's transition pattern. As for why some users with more than 300 records do not have a better performance, it is because there are few active users in the NYC dataset so the results can not reflect the actual performance at the group level. 

To measure the complexity of users' historical trajectories, we design two indicators, 'location entropy' and 'category entropy', for historical trajectories.
Assume that a set of user's visited locations is $L_u = [l_1,l_2,\cdots, l_k]$ and $\forall l_i\in L_u$,  $\eta_i$ is the number of times the user visited the specific location $l_i$. The location entropy $E_{loc}$ for this user can then be calculated as: 
\begin{align}
q_i & = \eta_i / \sum_{l_j\in L_u} \eta_j \\
E_{loc} & = -\sum_{l_i\in L_u}{ q_i \log(q_i) } 
\end{align}
Category entropy $E_{cat}$ can be calculated in a similar manner on the user's historical access category collection.

\begin{figure}[!htbp]
  \centering
  \includegraphics[width=0.98\linewidth]{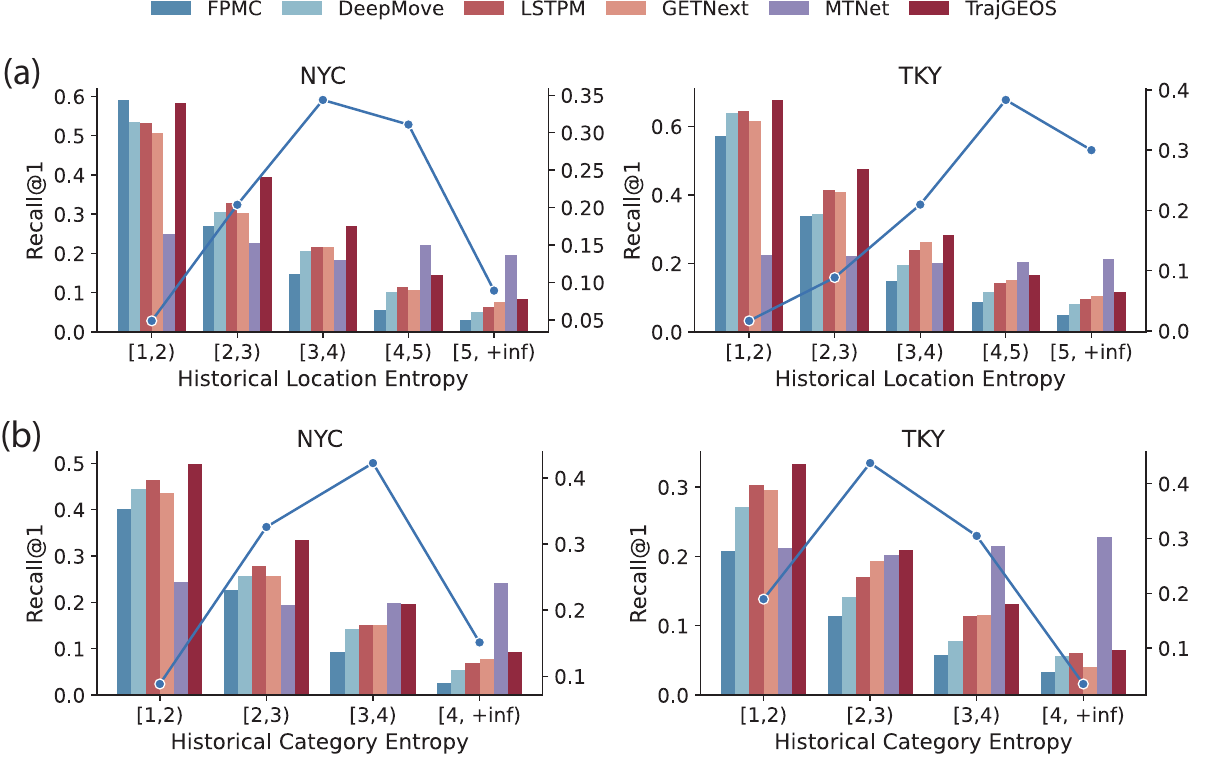}
  \caption{(a) Relationship of historical location complexity and predicted accuracy. (b) Relationship of historical category complexity and predicted accuracy. 
%  For each sub-figure, the y-axis on the left corresponds to the metric of prediction, which represents Recall@1. The y-axis on the right corresponds to the polyline, which indicates the proportion of users in the corresponding range of the x-axis in the test data of TrajGEOS.
  }
  \label{fig:batchEntropy_bar}
\end{figure}

To compare the performance of the models on users with different complexities, we divide the users into groups according to their location entropy and then calculate the average prediction performance of different models for each user group. The result is visualized by barplot in Figure~\ref{fig:batchEntropy_bar} (a). The polyline in Figure~\ref{fig:batchEntropy_bar} (a) corresponds to the y-axis on the right side and represents the proportion of the users belonging to different location entropy intervals in the test data for TrajGEOS. 

The category of visiting location is closely related to the user's activity, which reflects some internal logic of the user's mobility behavior. So we analyzed the impact of historical category complexities on prediction accuracy. Following the manner of Figure~\ref{fig:batchEntropy_bar} (a), we plot the prediction accuracies under different category entropies as shown in Figure~\ref{fig:batchEntropy_bar} (b). The results show that the more complex a user's historical categories are, the more difficult it is to accurately predict his next movement.

% TODO Move sensitivity analysis section to appendix

\section{Conclusion}

In this paper, we propose TrajGEOS, a trajectory graph enhanced orientation-based sequential model, to predict users' next locations. To capture the relationships among locations and a user's long-term preferences, we apply graph learning technology to the global trajectory graph.
Also, to explore a user's recent tendency, we design an orientation module that integrates position embedding techniques with attention mechanisms to compute their mid-term preferences. 
Moreover, our TrajGEOS model leverages a multi-task prediction strategy, where the next category prediction serves as an auxiliary task to enhance the next location prediction.
A series of experiments on three LBSN datasets demonstrate that our proposed model outperforms all state-of-the-art models, and ablation studies also confirm the effectiveness of our model's different components. 
For future work, we will verify the scalability of TrajGEOS on dense mobility datasets such as population-scale mobile phone data, and investigate whether friendship networks can contribute to the next location prediction tasks on LBSN datasets.

\section*{Acknowledgments}
This work was supported by the National Science Foundation of China (62102258), the Shanghai Municipal Science and Technology Major Project (2021SHZDZX0102) and the Fundamental Research Funds for the Central Universities. The computations in this paper were run on the AI for Science Platform supported by the Artificial Intelligence Institute at Shanghai Jiao Tong University.

% {
% \appendix[Proof of the Zonklar Equations]
% Use $\backslash${\tt{appendix}} if you have a single appendix:
% Do not use $\backslash${\tt{section}} anymore after $\backslash${\tt{appendix}}, only $\backslash${\tt{section*}}.
% If you have multiple appendixes use $\backslash${\tt{appendices}} then use $\backslash${\tt{section}} to start each appendix.
% You must declare a $\backslash${\tt{section}} before using any $\backslash${\tt{subsection}} or using $\backslash${\tt{label}} ($\backslash${\tt{appendices}} by itself
%  starts a section numbered zero.)
%  }

{\appendices
\section{Detailed algorithm of graph learning module}

\begin{center}
%\begin{minipage}{0.9\linewidth}
\begin{algorithm}[!htbp]
    \BlankLine
    \SetKwInOut{Input}{Input}
    \SetKwInOut{Output}{Output}
    \caption{Graph learning}
    \label{alg:graph_learning}
    \Input{Trajectory graph $G$ with node set $V$, edge set $E$, 
        location set $L = \{l_i\}_{i\leq M}$. User set $U$. 
        }
    \Output{Location embeddings $\{z_i\,|\, \forall v_i \in V\}$ ; 
       User embeddings $\{u_i^{long} \,|\, \forall u_i \in U \}$;}
    \BlankLine
    \For{$v_i \in V$}{ 
            $h_i^0 \leftarrow$  initialize node according to Eq.~\ref{eq:init_h0} \;
    }
    \For{$e_{i \rightarrow j} \in E$}{
        $\mathcal{E}_{i\rightarrow j}^0 \leftarrow$ initialize edge according to Eq.~\ref{eq:init_edge} \;
    }
    \For{$k=1,2$}{
        Update $\{ h_{i}^k \;|\; \forall v_i \in V \}$ according to Eq.~\ref{eq:global_node_update1}~\ref{eq:global_node_update2} \;
        Update $\{ \mathcal{E}_{i\rightarrow j}^k \;|\; \forall e_{i \rightarrow j} \in E \}$ according to Eq.~\ref{eq:global_edge_update} \;
        }
    $z_i \leftarrow$  $\forall v_i \in V$ calculate $z_i$ according to Eq.~\ref{eq:fusion} \;
    \For {$u_i \in U$} { 
        $G_i \leftarrow$  the trajectory graph of $u_i$ \;
        $V_i \leftarrow$ node set in $G_i$ \;
        $h_{i,j}^0 \leftarrow z_j\;,\; \forall v_j \in V_i$ \; 
        \For{$k=1,2$}{
            Update $h_{i, j}^k$ according to Eq.~\ref{eq:subgraph_conv1}~\ref{eq:subgraph_conv2}, $\forall v_j\in V_i$ \;
            $h_{i, j}^k \leftarrow h_{i, j}^k / ||h_{i, j}^k||_2 \;,\; \forall v_j\in V_i$ \;
        }
        $u_i^{long} \leftarrow$ Calculate $u_i^{long}$ according to Eq.~\ref{eq:user_readout} \;
    }
	\Return {$\{z_i\,|\, \forall v_i \in V\}$, $\{u_i^{long}\,|\, \forall u_i \in U \}$}
\end{algorithm}
%\end{minipage}
\end{center}

\section{Detailed experimental parameters}

We implement our TrajGEOS model using the Pytorch framework and Pytorch Geometric library. The dimensions of user id embedding, category id embedding, weekday embedding, and hour embedding are set to 64, 64, 16, and 16 for all datasets. The dimension of the hidden state in all GRUs is 256 and $\alpha$ for multi-task learning and is set to 0.7.

\textbf{Settings in graph learning.} We use location ID, category and GPS coordinates information to initialize the embedding for nodes in the trajectory graph. 
We set two 64-dim embedding layers to encode raw location ID and category data. Then we concatenate location embedding, category embedding, and the original latitude and longitude to generate the initial embedding for nodes.
The dimension of nodes in the trajectory graph, which corresponds to the actual location, is set to 128, And the readout of each user's trajectory graph is a 128-dimensional embedding. 
When it comes to graph convolution operation, we set the edge dropout ratio to $0.5$ and add dropout layers with $p=0.5$ between graph convolution layers in both the lower convolution on the global trajectory graph and the higher convolution on the user's trajectory graph. 
The fusion module within the global graph convolution employs a linear layer that takes both the original location embedding and the location embedding obtained after graph convolution as input. The output embedding dimension is identical to that of the global graph convolution feature. 

\textbf{Optimizer and learning rate.} TrajGEOS uses the Adam optimizer on all these 3 datasets. We set the learning rate to $1e-4$ for all three datasets and set the weight decay parameter to $1e-4$. More experimental settings can be found in our code.

\section{Additional experimental results}

\subsection{Next category prediction results}

Table~\ref{tab:cat_result} shows the predicted category prediction results of our model and baseline models, using the same experimental setup as Table~\ref{tab:main_result}. 
Table~\ref{tab:main_result} has proved that TrajGEOS outperforms all baseline models on the next location prediction task, yet it does not exceed all baseline models regarding next category prediction. 

\begin{table}[!htbp]
\centering
\caption{Results of next category prediction task}
\label{tab:cat_result}
\resizebox{\linewidth}{!}{
\begin{threeparttable} 
\begin{tabular}{ccccc|cccc} 
\hline
\multirow{2}{*}{Model} & \multicolumn{4}{c|}{NYC}                        & \multicolumn{4}{c}{TKY}                         \\ 
\cline{2-9}
                       & R@1        & R@5 & R@10 & M@10 & R@1        & R@5 & R@10 & M@10  \\ 
\hline
FPMC-D                 & 0.2191          & 0.5547   & 0.6863    & 0.3597 & 0.4292          & 0.6208   & 0.7119   & 0.5119  \\
FPMC-W                 & 0.2051          & 0.5419   & 0.6816    & 0.3456 & 0.4217          & 0.6085   & 0.7030   & 0.5034  \\
DeepMove               & 0.2470          & 0.4956   & 0.5767    & 0.3527 & 0.3960          & 0.5628   & 0.6227   & 0.4687  \\
GeoSAN                 & 0.1738          & 0.4585   & 0.6304    & 0.2970 & 0.2543          & 0.5629   & 0.7188   & 0.3873  \\
% Flashback$^*$             & 0.2804          & 0.6213   & 0.7403    & 0.4233 & \textbf{0.4795}          & \textbf{0.7298}   & \textbf{0.8017}   & \textbf{0.5851}  \\
LSTPM                 & 0.2965          & 0.6386   & 0.7538    & 0.4408 & 0.4492          & 0.7104   & 0.7874   & 0.5606  \\
GETNext                & \underline{0.3030}          & \underline{0.6458}   & \underline{0.7561}    & \underline{0.4486} & 0.4457          & 0.7610   & 0.8457   & 0.5802  \\
MTNet           & 0.2679 & 0.5594 & 0.6692 & 0.3943 & 0.4577 & 0.7321 & 0.8202 & 0.5762 \\ 
TrajGEOS               & \textbf{0.3329}    & \textbf{0.6746}  & \textbf{0.7755}    & \textbf{0.4780}   & \underline{0.4779}   & \underline{0.7201}   & \underline{0.7845}   & \underline{0.5812}     \\ 
\hline
TrajGEOS-woGraph       & 0.3218          & 0.6584   & 0.7571    & 0.4643 & 0.4683          & 0.7011   & 0.7659   & 0.5678  \\
TrajGEOS-onlyGraph     & 0.3249          & 0.6571   & 0.7560    & 0.4660 & 0.4755          & 0.7138   & 0.7785   & 0.5773  \\
TrajGEOS-woShort       & 0.3150          & 0.6625   & 0.7688    & 0.4605 & 0.4650          & 0.7065   & 0.7750   & 0.5681  \\
TrajGEOS-woMid         & 0.3155          & 0.6600   & 0.7643    & 0.4610 & 0.4749          & 0.7163   & 0.7831   & 0.5780  \\ 
\hline
\end{tabular}
% \begin{tablenotes}    %这行要添加， 从这开始
%     \footnotesize               %这行要添加
%     \item[*] Flashback adopts stricter policies to filter more sparse users during data preparation, reducing the difficulty of prediction.
% \end{tablenotes} 
\end{threeparttable}
}
\end{table}

To further analyze this issue, we increase the number of epochs for TrajGEOS on different datasets and find that the next category prediction task reaches peak performance later than the next location prediction task. 
Although increasing the number of training epochs can allow TrajGEOS to achieve better results on the next category prediction tasks, it will affect the performance of the next location prediction. 
Since next location prediction is the main task of this paper, we put priority on optimizing the location prediction task when setting parameters.

\subsection{Performance on different types of locations}

\begin{figure}[!htbp]
  \centering
  \includegraphics[width=\linewidth]{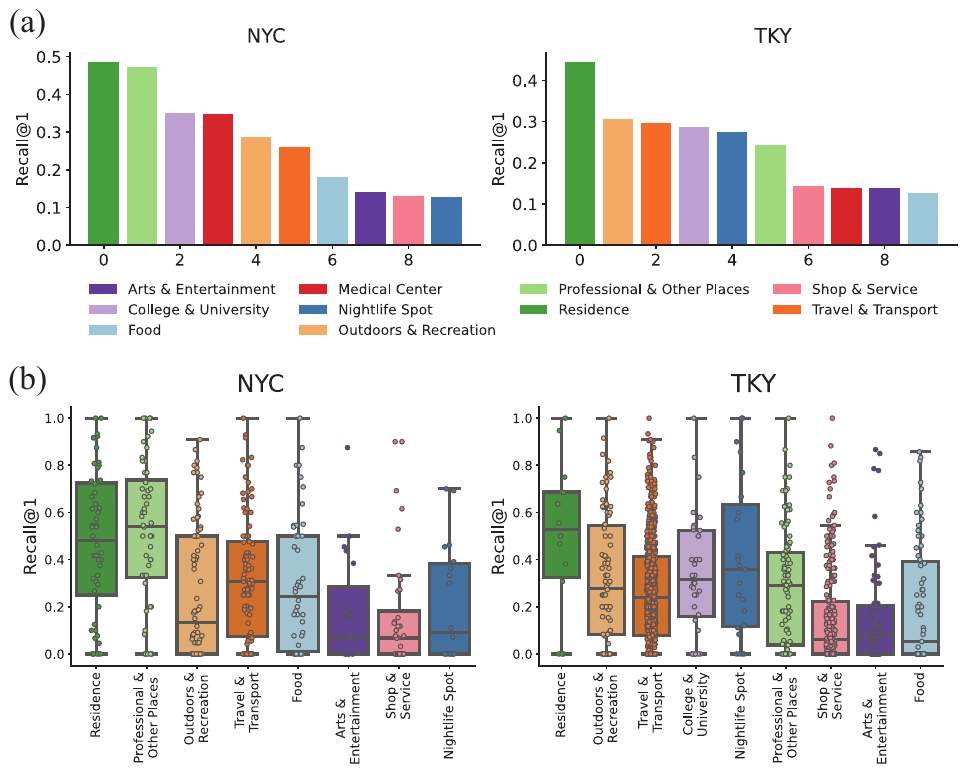}
  \caption{(a) The average performance of different super-categories. (b) The predicted accuracies of locations in different super-categories.}
  \label{fig:supCat_batchPlot}
\end{figure}

We check the prediction accuracy for locations that belong to different super-categories in the Foursquare dataset. We first calculate the average prediction accuracies for each super-category and plot them in Figure~\ref{fig:supCat_batchPlot} (a). After calculating the prediction accuracies for every location and categorizing the locations by their super-category, we draw boxplots and strip charts for every super-category as Figure~\ref{fig:supCat_batchPlot} (b) shows. 
From Figure~\ref{fig:supCat_batchPlot} we find that locations in the `Residence' category are easier to predict. And there are a large number of `Travel \& Transport' locations in the TKY dataset, such as `Subway', `Train station', etc., and the processing of this kind of location can be future research content. 

\subsection{Sensitivity Analysis}

\begin{table}[!htbp]
\centering
\caption{Results of different $\alpha$ in multi-task learning}
\label{tab:alpha_parameter}
\resizebox{\linewidth}{!}{
\begin{tabular}{ccccc|cccc} 
\hline
\multirow{2}{*}{$\alpha$} & \multicolumn{4}{c|}{Category}   & \multicolumn{4}{c}{Location}   \\ 
\cline{2-9}
    & R@1     & R@5     & R@10    & M@10    & R@1     & R@5     & R@10    & M@10     \\
\hline
0.1       & 0.3336 & 0.6782 & 0.7796 & 0.4796 & 0.2619 & 0.5584 & 0.6565 & 0.3857   \\
0.3       & 0.3348 & 0.6800 & 0.7810 & 0.4809 & 0.2684 & 0.5721 & 0.6692 & 0.3948   \\
0.5       & 0.3328 & 0.6773 & 0.7781 & 0.4788 & 0.2692 & 0.5733 & 0.6709 & 0.3963   \\
0.7       & 0.3328 & 0.6745 & 0.7755 & 0.4779 & 0.2721 & 0.5739 & 0.6715 & 0.3982   \\
0.9       & 0.3283 & 0.6673 & 0.7669 & 0.4722 & 0.2715 & 0.5744 & 0.6717 & 0.3981   \\
\hline
\end{tabular}
}
\end{table}

We analyze the influence of hyperparameters $\alpha$ in multi-task learning. 
Setting different $\alpha$ on the NYC dataset, we get different results like Table~\ref{tab:alpha_parameter} shows. Table~\ref{tab:alpha_parameter} shows we can observe that $\alpha$ has an impact on the final results of both category prediction and location prediction. 
Although the difference in results that have different $\alpha$ is not large, we choose $\alpha=0.7$ as our optimal parameter based on the evaluation metric $Recall@1$ for the next location prediction task.

\begin{table}[!htbp]
\centering
\caption{ The results of different lengths of user's recent trajectory on NYC dataset }
\label{tab:midLen_analysis}
\resizebox{\linewidth}{!}{
\begin{tabular}{ccccc|cccc} 
\hline
\multirow{2}{*}{\begin{tabular}[c]{@{}c@{}}Recent trajectory\\ length (week)\end{tabular}} & \multicolumn{4}{c|}{Category}             & \multicolumn{4}{c}{Location}               \\ 
\cline{2-9}
    & R@1 & R@5 & R@10 & M@10  & R@1 & R@5 & R@10 & M@10   \\ 
\hline
1     & 0.3268  & 0.6672  & 0.7687   & 0.4704 & 0.2682  & 0.5677  & 0.6646   & 0.3930  \\
2     & 0.3328  & 0.6745  & 0.7755   & 0.4779 & 0.2721  & 0.5739  & 0.6715   & 0.3982  \\
4     & 0.3313  & 0.6794  & 0.7780   & 0.4784 & 0.2712  & 0.5762  & 0.6732   & 0.3986  \\
6     & 0.3292  & 0.6783  & 0.7793   & 0.4769 & 0.2713  & 0.5763  & 0.6737   & 0.3987  \\
8     & 0.3245  & 0.6773  & 0.7777   & 0.4733 & 0.2675  & 0.5765  & 0.6748   & 0.3968  \\
\hline
\end{tabular}
}
\end{table}

In addition, we analyze the length of the recent historical trajectory used by our TrajGEOS and set the length of the recent historical trajectory used as 1, 2, 4, 6, and 8 weeks respectively. The prediction results on the NYC dataset are shown in Table~\ref{tab:midLen_analysis}. 
Based on the $Recall@1$ evaluation metric for the next location prediction task, we finally choose to use the user's historical trajectory from the past two weeks for both modeling and forecasting.

\subsection{The effectiveness of orientation module}

To explain why we use the Orientation module instead of directly using GRU to learn user mid-term preferences, we replace the Orientation module in TrajGEOS with GRU to process the recent trajectory information, and the output is used as the initial state to process the GRU of the current trajectory. 
The results obtained by setting different lengths of the recent sub-trajectories utilized to calculate the user's mid-term preference are shown in Table~\ref{tab:GRU_replace_orientation}.

\begin{table}[!htbp]
\centering
\caption{Replace the orientation module with GRU and check its performance on the NYC dataset with different lengths of sub-trajectories that compose the recent trajectories. Every recent sub-trajectory is the trace of a user in a week.}
\label{tab:GRU_replace_orientation}
\resizebox{\linewidth}{!}{
\begin{tabular}{ccccc|cccc} 
\hline
\multirow{2}{*}{\begin{tabular}[c]{@{}c@{}}Num of recent \\sub-trajectories\end{tabular}}   & \multicolumn{4}{c|}{Category}   & \multicolumn{4}{c}{Location}   \\ 
\cline{2-9}
    & R@1 & R@5 & R@10 & M@10 & R@1 & R@5 & R@10 & M@10  \\ 
\hline
1      & 0.3238   & 0.6670   & 0.7682    & 0.4683 & 0.2636   & 0.5697   & 0.6688    & 0.3913  \\
2      & 0.3257   & 0.6690   & 0.7708    & 0.4706 & 0.2655   & 0.5721   & 0.6714    & 0.3934  \\
4      & 0.3259   & 0.6702   & 0.7716    & 0.4707 & 0.2651   & 0.5714   & 0.6704    & 0.3925  \\
6      & 0.3272   & 0.6714   & 0.7728    & 0.4726 & 0.2667   & 0.5717   & 0.6703    & 0.3937  \\
8      & 0.3262   & 0.6711   & 0.7704    & 0.4712 & 0.2649   & 0.5709   & 0.6696    & 0.3920  \\
\hline
\end{tabular}
}
\end{table}

\subsection{Different number of EGraphSAGE layers on global trajectory graph}
To determine the optimal number of graph convolution layers on the global trajectory graph, we conducted experiments on the NYC dataset. 
We first set the number of convolution layers on the user trajectory subgraph to 2, and then tested 2, 3, 4, and 5 graph convolution layers on the global graph, respectively. 
The experimental results are shown in Table~\ref{tab:check_global_gnn_layers}. Based on the $Recall@1$ metric of the next place prediction task, we finally use 2-layer graph convolutions to generate the embedding of nodes on the global trajectory graph.

\begin{table}[!htbp]
\centering
\caption{Result on NYC dataset with different EGraphSAGE layer nums on global trajectory graph}
\label{tab:check_global_gnn_layers}
\resizebox{\linewidth}{!}{
\begin{tabular}{ccccc|cccc} 
\hline
\multirow{2}{*}{\begin{tabular}[c]{@{}c@{}}EGraphSAGE\\Layer Num\end{tabular}} & \multicolumn{4}{c|}{Category}            & \multicolumn{4}{c}{Location}              \\ 
\cline{2-9}
                           & R@1 & R@5 & R@10 & M@10 & R@1 & R@5 & R@10 & M@10  \\ 
\hline
2                          & 0.3329   & 0.6746   & 0.7755    & 0.4780 & 0.2721   & 0.5740   & 0.6716    & 0.3982  \\
3                          & 0.3324   & 0.6768   & 0.7765    & 0.4779 & 0.2709   & 0.5740   & 0.6709    & 0.3974  \\
4                          & 0.3314   & 0.6766   & 0.7743    & 0.4772 & 0.2704   & 0.5741   & 0.6710    & 0.3973  \\
5                          & 0.3326   & 0.6756   & 0.7752    & 0.4774 & 0.2715   & 0.5730   & 0.6708    & 0.3973  \\
\hline
\end{tabular}
}
\end{table}

\subsection{Different number of GraphSAGE layers on user's trajectory graph}
Under the condition of utilizing two global graph convolutional layers, we further investigate the impact of using different numbers of graph convolutional layers for subgraph links in user trajectories on experimental outcomes. 
We conduct experiments on the NYC dataset and present our findings in Table A. Based on the $Recall@1$ metric for the next place prediction task, we finally employ a 2-layer GraphSAGE for graph learning on user trajectory graphs.

\begin{table}[!htbp]
\centering
\caption{Result on NYC dataset with different GraphSAGE layer nums on user's trajectory graph}
\label{tab:check_sub_gnn_layers}
\resizebox{\linewidth}{!}{
\begin{tabular}{ccccc|cccc} 
\hline
\multirow{2}{*}{\begin{tabular}[c]{@{}c@{}}GraphSAGE\\Layer Num\end{tabular}} & \multicolumn{4}{c|}{Category}            & \multicolumn{4}{c}{Location}              \\ 
\cline{2-9}
                           & R@1 & R@5 & R@10 & M@10 & R@1 & R@5 & R@10 & M@10  \\ 
\hline
2                          & 0.3329   & 0.6746   & 0.7755    & 0.4780 & 0.2721   & 0.5740   & 0.6716    & 0.3982  \\
3                          & 0.3283   & 0.6680   & 0.7666    & 0.4725 & 0.2716   & 0.5722   & 0.6694    & 0.3975  \\
4                          & 0.3281   & 0.6673   & 0.7660    & 0.4722 & 0.2718   & 0.5724   & 0.6681    & 0.3971  \\
5                          & 0.3258   & 0.6617   & 0.7630    & 0.4686 & 0.2708   & 0.5720   & 0.6677    & 0.3963  \\
\hline
\end{tabular}
}
\end{table}

}

% \section{References Section}
% You can use a bibliography generated by BibTeX as a .bbl file.
%  BibTeX documentation can be easily obtained at:
%  http://mirror.ctan.org/biblio/bibtex/contrib/doc/
%  The IEEEtran BibTeX style support page is:
%  http://www.michaelshell.org/tex/ieeetran/bibtex/
 

\bibliographystyle{IEEEtran}
\bibliography{reference}

% \newpage

 

% \end{IEEEbiography}



% \vfill

\end{document}